\documentclass[11pt,a4paper]{article}

\usepackage[top=2.5cm, bottom=2.5cm, left=2.5cm, right=2.5cm]{geometry}

\usepackage{mathptmx}   

\usepackage{pifont} 
\usepackage{graphicx}
\usepackage{xcolor}
\usepackage{enumitem}
\usepackage{hyperref}
\usepackage{fancyhdr}
\usepackage{titlesec}
\usepackage{float}
\usepackage{ragged2e}
\usepackage{tabularx}
\usepackage{threeparttable}
\usepackage{setspace}
\usepackage[table]{xcolor} 
\usepackage{parskip} 
\usepackage{amsmath, amssymb}
\usepackage{booktabs}
\usepackage[numbers]{natbib}
\definecolor{teleai-blue}{RGB}{0, 80, 140}
\definecolor{teleai-lightblue}{RGB}{240, 250, 255}
\definecolor{darkred}{RGB}{220,20,60}    
\definecolor{darkblue}{RGB}{0,0,255}  
\definecolor{darkgreen}{RGB}{34,139,34}  
\definecolor{pureyellow}{RGB}{255,155,0}
\definecolor{titleblue}{RGB}{10,35,90}
\definecolor{checkgreen}{RGB}{0,160,60}
\definecolor{crossred}{RGB}{180,20,20}
\newcommand{\goodtick}{\textcolor{checkgreen}{\ding{51}}}
\newcommand{\badcross}{\textcolor{crossred}{\ding{55}}}

\titleformat{\section}
  {\sffamily\Large\bfseries}{\thesection}{1em}{}
\titleformat{\subsection}
  {\sffamily\large\bfseries}{\thesubsection}{1em}{}
\titleformat{\subsubsection}
  {\sffamily\normalsize\bfseries}{\thesubsubsection}{1em}{}

\usepackage{tcolorbox}
\tcbuselibrary{skins,breakable}

\newenvironment{customabstract}{
    \begin{tcolorbox}[
        colback=teleai-lightblue,
        colframe=teleai-lightblue,
        arc=3mm,
        boxrule=0pt,
        left=12pt,
        right=12pt,
        top=12pt,
        bottom=12pt,
        breakable
    ]
    \setstretch{1.15}
    \normalfont\normalsize
}{
    \end{tcolorbox}
    \vspace{-1.0em}
}

\makeatletter
\renewcommand{\@maketitle}{%
  \newpage
  \null
  \vskip 2em%
  \begin{flushleft}
    \let\footnote\thanks
    \LARGE\@title \par
    \vskip 1.5em
    \large\@author \par
    \vskip 1em
    \normalsize\@date \par
  \end{flushleft}
  \par
  \vskip 1.5em}
\makeatother

\begin{document}
\begin{customabstract}
\title{
    \vspace{-0.5cm}
    \parbox{\textwidth}{
        \RaggedRight
        \sffamily\LARGE\bfseries
        \setstretch{0.9}  
        ARAC: Benchmarking Auto-Research's Alignment and Completeness on End-to-End Researchs
    }\\[-0.9em]
}

\author{
    \RaggedRight
    \sffamily\large\bfseries 
    Jiale Cui, Yueyao Yuan, Kaixi Zhong, Xiaogang Xu, Jiafei Wu,    Zhe Liu\footnotemark[2] 
    \normalfont\normalsize
    \textbf{School of Software Technology, Zhejiang University} 
}
\date{}
\maketitle
\footnotetext[1]{Dataset: 
\href{https://github.com/cuijiale2004-hash/ARAC-Bench}{https://github.com/cuijiale2004-hash/ARAC-Bench}}
\thispagestyle{fancy} 

\vspace{-0.5cm}
The rapid advancement of Auto-Research has surfaced a fundamental evaluation challenge: \textit{how can we measure the alignment, logical coherence, and evolutionary completeness of its research trajectory with human research behavior}? We propose \textbf{A}uto-\textbf{R}esearch's \textbf{A}lignment and \textbf{C}ompleteness, \textbf{ARAC-Bench:} a \textbf{Researcher-Mimicking Evaluation } framework that shifts the objective from matching final answers to reproducing high-quality human research processes. The framework operates through two synergistic components: the \textbf{Academic Cognition Skills} system, which is the first to transforms implicit reviewer expertise into stage-calibrated, quantifiable rubrics; and a three-stage capability diagnostic protocol, which decomposes the research process under strict modular constraints into three traceable, mutually independent dimensions: \textbf{Proposal}, \textbf{Experiment}, and \textbf{Synthesis}. Systematic evaluation of 11 SOTA frameworks yields a best alignment score of only \textbf{67.9} of 100, revealing a significant gap in simulating rigorous human methodology. Validation against Ph.D. Candidates rankings shows a strong correlation of \textbf{0.8141}, confirming that ARAC-Bench reliably reflects the dimensions researchers truly value. ARAC-Bench provides not only a fine-grained diagnostic tool but also a scalable reward signal for training the next generation of autonomous research systems.

\end{customabstract}
\begin{figure}[htbp]   
    \centering
    \includegraphics[width=1\linewidth]{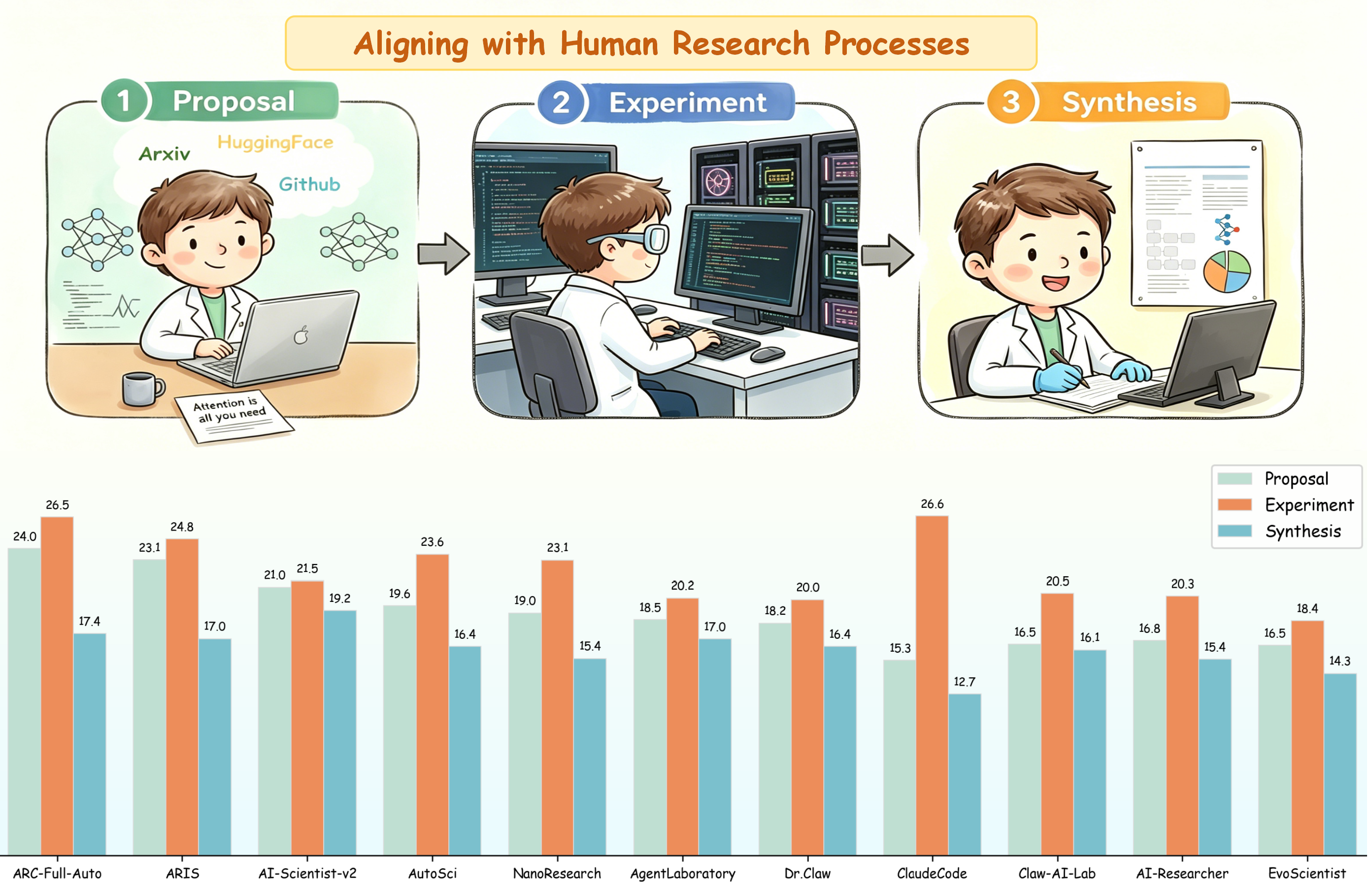}
    \label{fig:example}
\end{figure}

\definecolor{sectionblue}{RGB}{40,90,255} 
\titleformat{\section}
  {\color{sectionblue}\Large\bfseries}
  {\thesection}{0.5em}{}

\titleformat{\subsection}
  {\color{sectionblue}\large\bfseries}
  {\thesubsection}{0.5em}{}
  
\section{Introduction}

With the rapid development of large language models (LLMs) and autonomous agent technologies, Auto-Research has moved from conceptual exploration to preliminary practice. Frameworks such as ARIS \cite{yang2026aris}, Dr.Claw \cite{song2026drclaw}, and Claw-AI-Lab \cite{wu2026claw} have decomposed the research pipeline into distinct stages, enabling workflows encompassing ideation, experimentation, and synthesis to be completed autonomously \cite{kong2026ai,tang2026ai}. However, this progress has exposed a fundamental evaluation bottleneck: current paradigms cannot measure whether Agents truly emulate the cognitive processes of researchers. Existing approaches shown in Table \ref{tab:benchmarks11} fall into a dilemma: \textbf{automated scoring rubrics} rely on rigid scripts and sandbox execution, offering quantifiability but lacking semantic understanding of research quality; conversely, \textbf{LLM-as-Judge} can capture nuances yet suffers from unstructured subjective bias, hallucination, and poor reproducibility. 

More critically, these results-driven evaluations based on code pass rates or paper completeness \cite{tie2026autoresearch,wei2025ai,zheng2025automation} fail to distinguish genuine insights from brute-force search and overlook explorations that are theoretically sound but fail due to engineering coincidences \cite{beel2025evaluating,weng2025deepscientist,wan2026deep}. These metrics entirely bypass scrutiny of whether the research process adheres to human methodological standards, creating a dangerous disconnect between high scores and high-quality research \cite{tobias2025autonomous}.

\begin{table}[htbp]
\centering
\footnotesize
\renewcommand{\arraystretch}{1}
\begin{tabular}{
>{\centering\hspace{0pt}}m{2.8cm} 
>{\centering\arraybackslash}m{1cm} 
>{\centering\arraybackslash}m{0.1cm} 
>{\centering\arraybackslash}m{0.1cm} 
>{\centering\arraybackslash}m{0.1cm} 
>{\centering\arraybackslash}m{0.1cm} 
>{\centering\arraybackslash}m{4.6cm} 
>{\centering\arraybackslash}m{1.2cm} 
>{\centering\arraybackslash}m{1.9cm}
}
\toprule

\normalsize\textbf{Benchmark} & \footnotesize\textbf{Target} & \footnotesize\textbf{D} & \footnotesize\textbf{C} & \footnotesize\textbf{R} & \footnotesize\textbf{A} & \footnotesize\textbf{Metrics and Methods} & \footnotesize\textbf{Rules} & \footnotesize\textbf{Justification}\\

\midrule
\textbf{ResearchBench} & \cellcolor{pink!20}{LLM} & \textcolor{darkgreen}{\textbf{\textcolor{darkgreen}{\textbf{\checkmark}}}} & \textcolor{darkred}{\textbf{×}} & \textcolor{darkred}{\textbf{×}} & \textcolor{darkgreen}{\textbf{\checkmark}} & Recall, LLM & \textcolor{orange}{Static} & \cellcolor{violet!20}{Expert Review} \\
\arrayrulecolor{gray!40}
\midrule
\textbf{ScienceAgentBench} & \cellcolor{green!10}{Agent} & \textcolor{darkred}{\textbf{×}} & \textcolor{darkgreen}{\textbf{\checkmark}} & \textcolor{darkgreen}{\textbf{\checkmark}} & \textcolor{darkred}{\textbf{×}} & Execution, Accuracy, BERTScore & \textcolor{orange}{Static} & \cellcolor{violet!20}{Expert Review} \\
\midrule
\textbf{SciAgentArena} & \cellcolor{green!10}{Agent} & \textcolor{darkgreen}{\textbf{\checkmark}} & \textcolor{darkgreen}{\textbf{\checkmark}} & \textcolor{darkgreen}{\textbf{\checkmark}} & \textcolor{darkred}{\textbf{×}} & Execution, Accuracy & \textcolor{orange}{Static} & \cellcolor{yellow!20}{Replication} \\
\midrule
\textbf{EXP-Bench} & \cellcolor{green!10}{Agent} & \textcolor{darkgreen}{\textbf{\checkmark}} & \textcolor{darkgreen}{\textbf{\checkmark}} & \textcolor{darkgreen}{\textbf{\checkmark}} & \textcolor{darkgreen}{\textbf{\checkmark}} & Execution, Accuracy, LLM & \textcolor{orange}{Static} & \cellcolor{gray!20}None \\
\midrule
\textbf{MLRBench} & \cellcolor{pink!20}{LLM} & \textcolor{darkgreen}{\textbf{\checkmark}} & \textcolor{darkgreen}{\textbf{\checkmark}} & \textcolor{darkgreen}{\textbf{\checkmark}} & \textcolor{darkgreen}{\textbf{\checkmark}} & Execution, Accuracy, LLM  & \textcolor{orange}{Static} & \cellcolor{pink!20}Correlation \\
\midrule
\textbf{AstaBench} & \cellcolor{green!10}{Agent} & \textcolor{darkgreen}{\textbf{\checkmark}} & \textcolor{darkred}{\textbf{×}} & \textcolor{darkred}{\textbf{×}} & \textcolor{darkgreen}{\textbf{\checkmark}} & Accuracy, Recall, LLM & \textcolor{orange}{Static} & \cellcolor{pink!20}Correlation \\
\midrule
\textbf{MLAgentBench} & \cellcolor{pink!20}{LLM} & \textcolor{darkred}{\textbf{×}} & \textcolor{darkgreen}{\textbf{\checkmark}} & \textcolor{darkgreen}{\textbf{\checkmark}} & \textcolor{darkred}{\textbf{×}} & Execution, Accuracy & \textcolor{blue!60}{\textbf{Dynamic}} & \cellcolor{gray!20}None \\
\midrule
\textbf{MASSW} & \cellcolor{pink!20}{LLM} & \textcolor{darkgreen}{\textbf{\checkmark}} & \textcolor{darkred}{\textbf{×}} & \textcolor{darkred}{\textbf{×}} & \textcolor{darkgreen}{\textbf{\checkmark}} & BERTScore, ROUGE, BLEU & \textcolor{orange}{Static} & \cellcolor{pink!20}Correlation \\
\midrule
\textbf{AIRS-Bench} & \cellcolor{pink!20}{LLM} & \textcolor{darkgreen}{\textbf{\checkmark}} & \textcolor{darkgreen}{\textbf{\checkmark}} & \textcolor{darkgreen}{\textbf{\checkmark}} & \textcolor{darkred}{\textbf{×}} & Execution, Accuracy & \textcolor{orange}{Static} & \cellcolor{gray!20}None \\
\midrule
\textbf{ResearchCodeBench} & \cellcolor{pink!20}{LLM} & \textcolor{darkred}{\textbf{×}} & \textcolor{darkgreen}{\textbf{\checkmark}} & \textcolor{darkgreen}{\textbf{\checkmark}} & \textcolor{darkred}{\textbf{×}} & Execution, Accuracy & \textcolor{orange}{Static} & \cellcolor{gray!20}None \\
\midrule
\textbf{PaperBench} & \cellcolor{pink!20}{LLM} & \textcolor{darkred}{\textbf{×}} & \textcolor{darkgreen}{\textbf{\checkmark}} & \textcolor{darkgreen}{\textbf{\checkmark}} & \textcolor{darkgreen}{\textbf{\checkmark}} & Execution, Accuracy, LLM & \textcolor{blue!60}{\textbf{Dynamic}} & \cellcolor{yellow!15}{Replication} \\
\midrule
\textbf{CORE-Bench} & \cellcolor{green!10}{Agent} & \textcolor{darkred}{\textbf{×}} & \textcolor{darkgreen}{\textbf{\checkmark}} & \textcolor{darkgreen}{\textbf{\checkmark}} & \textcolor{darkred}{\textbf{×}} & Execution, Accuracy & \textcolor{orange}{Static} & \cellcolor{yellow!20}{Replication} \\
\midrule
\textbf{DSBENCH} & \cellcolor{pink!20}{LLM} & \textcolor{darkred}{\textbf{×}} & \textcolor{darkred}{\textbf{×}} & \textcolor{darkgreen}{\textbf{\checkmark}} & \textcolor{darkgreen}{\textbf{\checkmark}} & Accuracy, LLM & \textcolor{orange}{Static} & \cellcolor{violet!20}{Expert Review} \\
\midrule
\textbf{AInsteinBench} & \cellcolor{pink!20}{LLM} & \textcolor{darkred}{\textbf{×}} & \textcolor{darkgreen}{\textbf{\checkmark}} & \textcolor{darkgreen}{\textbf{\checkmark}} & \textcolor{darkred}{\textbf{×}} & Execution & \textcolor{blue!60}{\textbf{Dynamic}} & \cellcolor{violet!20}{Expert Review} \\
\midrule
\textbf{AstaBench} & \cellcolor{green!10}{Agent} & \textcolor{darkgreen}{\textbf{\checkmark}} & \textcolor{darkgreen}{\textbf{\checkmark}} & \textcolor{darkgreen}{\textbf{\checkmark}} & \textcolor{darkgreen}{\textbf{\checkmark}} & Recall, Accuracy, LLM & \textcolor{blue!60}{\textbf{Dynamic}} & \cellcolor{pink!20}Correlation \\
\arrayrulecolor{black}
\bottomrule
\end{tabular}
\caption{Comparison of existing research test sets. D: Design, C: Coding, R: Result, A: Analysis. In particular, we pay attention to whether the assessment rules are dynamically adjusted/adaptive, and whether there are valid Justification explanations to confirm their credibility.}
\label{tab:benchmarks11}
\end{table}

To address this issue, we propose \textbf{ARAC-Bench}, a Researcher-Mimicking Evaluation Framework, whose fundamental goal shifts from matching final answers to quantitative assessment of AI uto-Research's performance and gaps in simulating rigorous human research process. Unlike prior work, ARAC-Bench takes accepted papers from top conferences as the gold-standard reference and distills implicit reviewer expertise into \textbf{Academic Cognition Skills (ACS)}, a set of stage-aware, quantifiable scoring rubrics. ACS is not constructed from scratch; rather, it is derived from core scientific questions extracted from 7,000 AI papers accepted at NeurIPS, ICLR, and ICML over the past two years, along with cognitive strategies employed by authors and reviewers during rebuttals and discussions, which are then cleaned, polished, and structured. The wide range of sources enables ACS to possess cross-task and cross-domain transferability as well as good generalization capabilities, and it can be applied to different research directions of AI.

This design yields two salient advantages: (1) \textbf{Dual Human Alignment}: On one hand, ACS is derived from the review and discussion of real human experts, ensuring that the evaluation method aligns with the source of human cognition. On the other hand, the evaluation objective of ARAC-Bench is to measure whether the Auto-Research framework replicates human research methodology at the process level. This dual alignment not only provides a quantifiable basis but also ensures the semantic and cognitive validity of the evaluation.; (2) \textbf{Quantifiability}: it transforms subjective research quality into structured, measurable indicators at different stages, bridging the gap between the rigor of automated scoring rubrics and the semantic depth of LLM judges.

Concretely, ARAC-Bench is built upon structured annotations of 200 accepted papers from ICLR 2026 and decomposes the research workflow into three progressive stages: Proposal, Experiment, and Synthesis. Each stage is evaluated under a unified controlled-variable Researcher-Mimicking Evaluation, with all Auto-Research sharing the same base model, Kimi, and the ACS knowledge base. To ensure the fairness of the assessment and prevent cheating, we strictly limit the literature search time range to mid-2025, physically preventing the model from directly accessing the real papers serving as Gold References; in the Proposal stage, related-work scoring uses the cited reference set of the original paper as Ground Truth; in the Experiment stage, code implementation scoring is based on a predefined standard module library, with missing modules scored as 0, reinforcing engineering completeness constraints. \textit{Moreover, as AI technologies rapidly evolve, ARAC-Bench will dynamically update its leaderboard and continuously add new Gold References to ensure long-term validation.}

Validation against PhD-level expert rankings yields a strong average correlation of \textbf{0.8141}, confirming that ARAC-Bench, while remaining computationally feasible, reliably reflects the dimensions researchers truly value. Systematic evaluation of frameworks reveals that even the best-performing system achieves only \textbf{67.9} alignment and completeness, exposing a significant gap in simulating rigorous human methodology.

We summarize our contributions as follows:

\begin{itemize}
    \item \textbf{Researcher-Mimicking Evaluation Framework:} We propose ARAC-Bench, the first benchmark evaluating Auto-Research by reproducing \textbf{high-quality human research processes}. Using top-tier papers as Gold References, it simultaneously achieves alignment with human judgment, computational quantifiability, and continuous evolution capability.
    \item \textbf{ACS as Quantifiable Rubrics:} We distill transferable academic cognition skills from papers and their rebuttals, refining implicit expert cognition into stage-calibrated, measurable structured signals. ARAC-Bench transforms subjective research standards into objective indicators naturally aligned with LLM reasoning.
    \item \textbf{Fine-grained Alignment Diagnosis:} Through stage-wise evaluation and variable isolation across frameworks, we provide granular measurements of where AI deviates from human research cognition. The strong correlation with expert rankings validates ARAC-Bench as a reliable proxy for human judgment.
\end{itemize}

\section{Related Work}
\subsection{Autonomous Scientific Research Systems}

With the iterative advancement of LLMs, autonomous scientific research systems have transitioned from conceptual frameworks to practical implementations. Early systems, such as Co-scientist \cite{gottweis2025towards,gottweis2026accelerating}, demonstrated the feasibility of LLMs in manipulating laboratory equipment, while the AI Scientist series achieved end-to-end automation from hypothesis generation to manuscript drafting. However, these single-agent systems lack adaptive error correction and cross-task knowledge accumulation. Similarly, while Agent Laboratory  \cite{schmidgall2025agent}has automated specific research phases, it lacks ground-truth verification and experience distillation. In the domain of fully automated pipelines, ARIS \cite{yang2026aris} achieves closed-loop error correction through a dual-model adversarial mechanism, AutoSOTA \cite{li2026autosota} focuses on model reproduction via multi-agent collaboration, and EvoScientist \cite{lyu2026evoscientist} realizes cross-task self-evolution using dual persistent memory modules. As shown in Table \ref{tab:framework_stepwise_optimization}, despite the establishment of a comprehensive technical framework, existing systems still exhibit notable deficiencies in result validation, exception handling, experimental reproducibility, and the construction of rigorous scientific closed loops.

\begin{table}[htbp]
\centering
\begin{tabular}{lcccccccccc}
\toprule
\small{\textbf{Framework}} & \footnotesize{\textbf{Search}} & \footnotesize{\textbf{Design}} & \footnotesize{\textbf{Coding}} & \footnotesize{\textbf{Iteration}} & \footnotesize{\textbf{Memory}} & \footnotesize{\textbf{Analysis}} & \footnotesize{\textbf{Writing}} & \footnotesize{\textbf{Evolve}} & \footnotesize{\textbf{Review}} \\
\midrule

\small{ARC-Full-Auto\cite{liu2026autoresearchclaw}} & \textcolor{darkgreen}{$\checkmark$} & \textcolor{darkgreen}{$\checkmark$} & \textcolor{darkgreen}{$\checkmark$} & \textcolor{darkgreen}{$\checkmark$} & \textcolor{darkgreen}{$\checkmark$} & \textcolor{darkgreen}{$\checkmark$} & \textcolor{darkgreen}{$\checkmark$} & \textcolor{darkgreen}{$\checkmark$} & \textcolor{darkgreen}{$\checkmark$} \\
\small{Dr. Claw\cite{song2026drclaw}}           & \textcolor{darkgreen}{$\checkmark$} & \textcolor{darkgreen}{$\checkmark$} & \textcolor{darkgreen}{$\checkmark$} & \textcolor{darkgreen}{$\checkmark$} & \textcolor{darkred}{$\circ$} & \textcolor{darkgreen}{$\checkmark$} & \textcolor{darkgreen}{$\checkmark$} & \textcolor{darkred}{$\circ$} & \textcolor{darkgreen}{$\checkmark$} \\
\small{Claw-AI-Lab\cite{wu2026claw}}      & \textcolor{darkgreen}{$\checkmark$} & \textcolor{darkgreen}{$\checkmark$} & \textcolor{darkgreen}{$\checkmark$} & \textcolor{darkgreen}{$\checkmark$} & \textcolor{darkgreen}{$\checkmark$} & \textcolor{darkgreen}{$\checkmark$} & \textcolor{darkgreen}{$\checkmark$} & \textcolor{darkgreen}{$\checkmark$} & \textcolor{darkgreen}{$\checkmark$} \\
\small{AI-Scientist-v2\cite{yamada2025ai}}    & \textcolor{darkgreen}{$\checkmark$} & \textcolor{darkgreen}{$\checkmark$} & \textcolor{darkgreen}{$\checkmark$} & \textcolor{darkgreen}{$\checkmark$} & \textcolor{darkred}{$\circ$} & \textcolor{darkgreen}{$\checkmark$} & \textcolor{darkgreen}{$\checkmark$} & \textcolor{darkred}{$\circ$} & \textcolor{darkgreen}{$\checkmark$} \\
\small{NanoResearch\cite{xu2026nanoresearch}}       & \textcolor{darkgreen}{$\checkmark$} & \textcolor{darkgreen}{$\checkmark$} & \textcolor{darkgreen}{$\checkmark$} & \textcolor{darkred}{$\circ$} & \textcolor{darkred}{$\circ$} & \textcolor{darkgreen}{$\checkmark$} & \textcolor{darkgreen}{$\checkmark$} & \textcolor{darkred}{$\circ$} & \textcolor{darkred}{$\circ$} \\
\small{EvoScientist\cite{lyu2026evoscientist}}       & \textcolor{darkred}{$\circ$} & \textcolor{darkgreen}{$\checkmark$} & \textcolor{darkgreen}{$\checkmark$} & \textcolor{darkgreen}{$\checkmark$} & \textcolor{darkgreen}{$\checkmark$} & \textcolor{darkred}{$\circ$} & \textcolor{darkgreen}{$\checkmark$} & \textcolor{darkgreen}{$\checkmark$} & \textcolor{darkred}{$\circ$} \\
\small{AI-Researcher\cite{tang2026ai}}    & \textcolor{darkgreen}{$\checkmark$} & \textcolor{darkgreen}{$\checkmark$} & \textcolor{darkgreen}{$\checkmark$} & \textcolor{darkgreen}{$\checkmark$} & \textcolor{darkred}{$\circ$} & \textcolor{darkgreen}{$\checkmark$} & \textcolor{darkgreen}{$\checkmark$} & \textcolor{darkred}{$\circ$} & \textcolor{darkred}{$\circ$} \\
\small{ARIS\cite{yang2026aris}}               & \textcolor{darkgreen}{$\checkmark$} & \textcolor{darkgreen}{$\checkmark$} & \textcolor{darkgreen}{$\checkmark$} & \textcolor{darkgreen}{$\checkmark$} & \textcolor{darkred}{$\circ$} & \textcolor{darkgreen}{$\checkmark$} & \textcolor{darkgreen}{$\checkmark$} & \textcolor{darkred}{$\circ$} & \textcolor{darkgreen}{$\checkmark$} \\
\small{AutoSci\cite{qian2026autosci}}    & \textcolor{darkgreen}{$\checkmark$} & \textcolor{darkgreen}{$\checkmark$} & \textcolor{darkgreen}{$\checkmark$} & \textcolor{darkgreen}{$\checkmark$} & \textcolor{darkred}{$\circ$} & \textcolor{darkgreen}{$\checkmark$} & \textcolor{darkgreen}{$\checkmark$} & \textcolor{darkred}{$\circ$} & \textcolor{darkred}{$\circ$} \\
\small{AgentLaboratory\cite{schmidgall2025agent}}               & \textcolor{darkgreen}{$\checkmark$} & \textcolor{darkgreen}{$\checkmark$} & \textcolor{darkgreen}{$\checkmark$} & \textcolor{darkgreen}{$\checkmark$} & \textcolor{darkred}{$\circ$} & \textcolor{darkgreen}{$\checkmark$} & \textcolor{darkgreen}{$\checkmark$} & \textcolor{darkred}{$\circ$} & \textcolor{darkred}{$\circ$} \\
\bottomrule
\end{tabular}
\caption{Whether a framework has added unique design features or optimization to better support the Auto-Research where \textit{optimization} specifically refers to whether, beyond implementing core functionalities, they have added distinctive design features to better support automation in research tasks.
}
\label{tab:framework_stepwise_optimization}
\end{table}

\subsection{Benchmarks for Scientific AI}

Concurrently, benchmarks for scientific AI have been continuously refined. ScienceAgentBench \cite{chen2025scienceagentbench} and PaperBench \cite{starace2025paperbench} quantify capabilities in scientific discovery and experimental implementation based on peer-reviewed publications and hierarchical evaluation frameworks, respectively, while CORE-Bench and ReproduceBench \cite{zhao2025autoreproduce} focus on the computational reproducibility of code and data. Frontier benchmarks have further expanded the evaluation dimensions: ResearchBench \cite{liu2025researchbench}, AIRS-Bench \cite{yang2026aris}, and FIRE-Bench \cite{wang2026fire} assess the capacity for full-process rediscovery in complex tasks and cutting-edge scenarios, and SGI-Bench explores generalized evaluation paradigms. Furthermore, AutoResearchBench \cite{xiong2026autoresearchbench} concentrates on the preliminary literature review phase, specifically targeting literature retrieval and academic synthesis capabilities, whereas ARC-Bench \cite{liu2026autoresearchclaw} covers the entire pipeline from research to writing. Nevertheless, current evaluations generally suffer from limitations like Table \ref{tab:benchmarks11}: they predominantly focus on outcome metrics such as task completion rates, neglecting normative constraints throughout the research process and lacking procedural assessments of adherence to rigorous scientific methodologies and academic logic \cite{wan2026deep}.

\subsection{LLM-Assisted Automated Peer Review}
Another critical research trajectory involves LLM-assisted automated peer review. Early studies merely achieved superficial optimization, whereas subsequent works have generated substantive critiques leveraging public datasets. OpenReviewer \cite{idahl2025openreviewer} and DeepReview \cite{zhu2025deepreview} have enhanced rigor through dimensional structured reviews and expert-level deep reasoning modes, respectively, while CycleReviewer \cite{weng2025cycleresearcher} uncovers flaws via adversarial iterations. Multi-agent frameworks have further elevated review quality; for instance, ScholarPeer \cite{goyal2026scholarpeer} reconstructs the logic of senior experts through role-based collaboration, and ReviewAgents \cite{gao2025reviewagents} integrate chain-of-thought reasoning with literature retrieval to generate refined feedback. However, optimization solely targets the quality of final comments without structurally modeling the cognitive reasoning process of the review, thereby failing to provide traceable and diagnosable fine-grained feedback \cite{weng2026deepreviewer}, further weakening the rigor of the Benchmark.

\section{ARAC-Bench}

The design of ARAC-Bench revolves around a central proposition: how can the research process of automated research frameworks be evaluated in a manner that is both aligned with human expert cognition and computationally quantifiable?

\subsection{Academic Cognition Skills}
To ensure credible human-aligned evaluation, we eschew generic LLM prompts and public rebuttals, which lack structured quantification or omit consensus-based methodological principles that have reached strong consensus and thus remain unchallenged. To this end, we propose \textbf{Academic Cognition Skills}, employing a \textbf{Reviewer Skill Distillation} approach that conceptualizes evaluation criteria as transferable cognitive skills demonstrated by expert reviewers in top-tier publications, rather than static rules.

\begin{figure}[htbp]   
    \centering
    \includegraphics[width=0.88\linewidth]{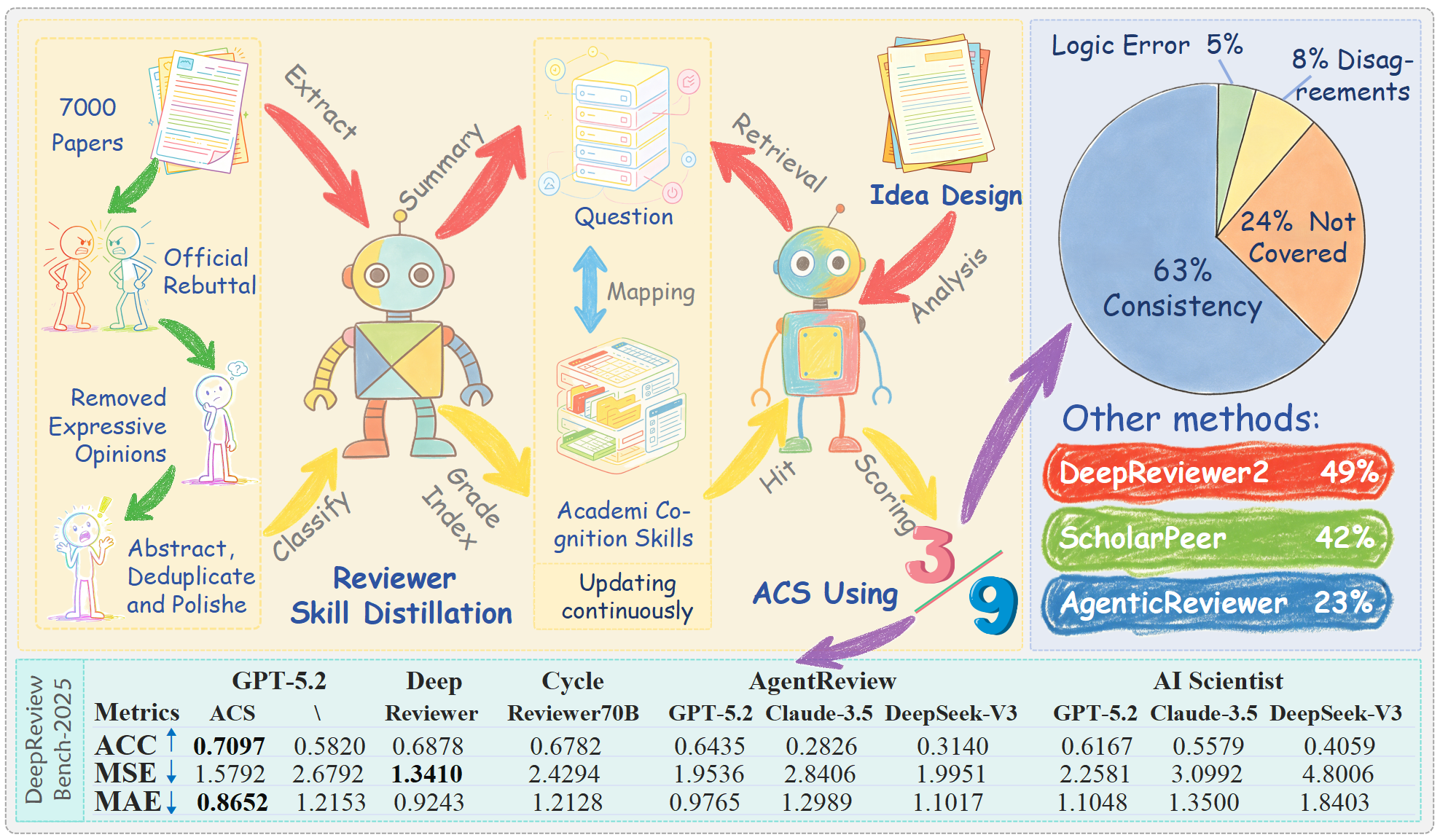}
    \vspace{-1em}
    \caption{To validate its reliability, we evaluated ACS on 200 paper queries of unseen ICLR papers above, achieving the SOTA 63\% consistency rate. Furthermore, we tested ACS on DeepReviewBench-2025 in the bottom of Figure and found that ACS was able to significantly improve the accuracy of the scores. Since ACS is based on data from 2025-2026, we found the results were not satisfactory in Bench-2024, which also compelled us to decide to regularly update the skills.}
    \label{fig:Drawing_05}
\end{figure}

The construction of ACS originates from deep mining of large-scale, high-quality academic corpora. From 7,000 top AI conference papers accepted at NeurIPS, ICLR, and ICML in 2025 and 2026, we first extracted the core scientific question each paper aimed to address; subsequently, we targeted the corresponding rebuttals and review discussions to extract the cognitive strategies employed by authors and reviewers in defending, clarifying, or revising those questions. These strategies were cleaned, deduplicated, and polished by experts, and then refined into structured ACS entries. This process ensures that ACS covers not only explicit points of contention but also implicit domain consensus and rigorous research structures. The resulting ACS knowledge base encompasses five major themes: (1)Large Language Models, (2)Multimodal Learning, (3)Diffusion Models, (4)Reinforcement Learning, and (5)Deep Learning, and 121 sub-themes.

As shown in Figure \ref{fig:Drawing_05}, ACS evaluation follows a progressive operational pathway: it begins with a comprehensive analysis of the research problem's context, from which a set of highly relevant core questions is extracted. These questions are then systematically matched against a predefined library of skill standards. 

\begin{tcolorbox}[
    colback=white,
    colframe=titleblue,
    arc=3pt, 
    boxrule=1pt,
    title={Example of Agentic RL for Tools in Long-Horizon tasks...},
    colbacktitle=titleblue,
    coltitle=white,
    fonttitle=\bfseries\Large,
    top=4pt, bottom=4pt, left=4pt, right=4pt
]
\begin{flushleft}
\par\vspace{3pt}
\noindent\goodtick\quad\textbf{Clarify the applicability of MDP and POMDP}(\textit{If Completely Considered})\\
\end{flushleft}

The research must clearly define whether the task is fully observable (MDP) or partially observable (POMDP). For Long-Horizon tasks, since the environmental state accumulates uncertainty over time and the agent's perception is often limited, POMDP is usually a more accurate theoretical abstraction...

\par\vspace{3pt}
\noindent\badcross\quad\textbf{Sample Efficiency and Trajectory Quality} (\textit{If Not Considered Completely})\\
The Long-Horizon Agentic tasks face a fundamental bottleneck of scarce high-quality trajectories. The research plan should explain how it ensures trajectory quality under high sample efficiency - through selective sampling, experience replay optimization, or other mechanisms...

\par\vspace{3pt}
\noindent\goodtick\quad\textbf{Delayed Credit Assignment Solvability}(\textit{If Completely Considered})\\
In Long-Horizon, the final result rewards are often sparse, and the contributions of intermediate steps are difficult to trace. The research plan must clearly explain how it solves the credit assignment problem - whether through hierarchical relative advantage estimation, attribution of tool invocation advantages, or by using the data filtering wheel to convert sparse rewards into clear filtering signals...

\end{tcolorbox}

Based on the matching results, the top five most closely aligned skill standards above are selected and used as the primary basis for final assessment. This process ensures that evaluation starts from the specific research context while leveraging a standardized skill framework to achieve systematicity and traceability, enabling assessments that are both grounded in actual research progress and supported by structured quantitative criteria. This design naturally integrates the contextual adaptability of LLM-as-Judge with the structural quantifiability of a standardized scoring system, aligning seamlessly with human experts' cognitive pathways.

\subsection{Researcher-Mimicking: Capability Diagnosis of Proposal, Experiment, and Synthesis}

Building upon the ACS system, ARAC-Bench constructs a stage-wise capability diagnostic protocol like Figure \ref{fig:Three-Stage}. To ensure evaluation fairness and ecological validity, we enforce strict temporal isolation: the literature search and knowledge base access of all evaluated frameworks are hard-truncated to mid-2025, physically eliminating any possibility that a model could directly retrieve the real papers serving as Gold References or their subsequent citation networks via search. The Researcher-Mimicking Evaluation requires that a Auto-Research framework complete only one stage at a time, which ensures that each score can be precisely attributed to a specific ACS deficiency or engineering capability defect.
. The three stages sum to a total of 100, with the specific diagnostic protocols as follows:

\begin{figure}[htbp]   
    \centering
    \includegraphics[width=0.98\linewidth]{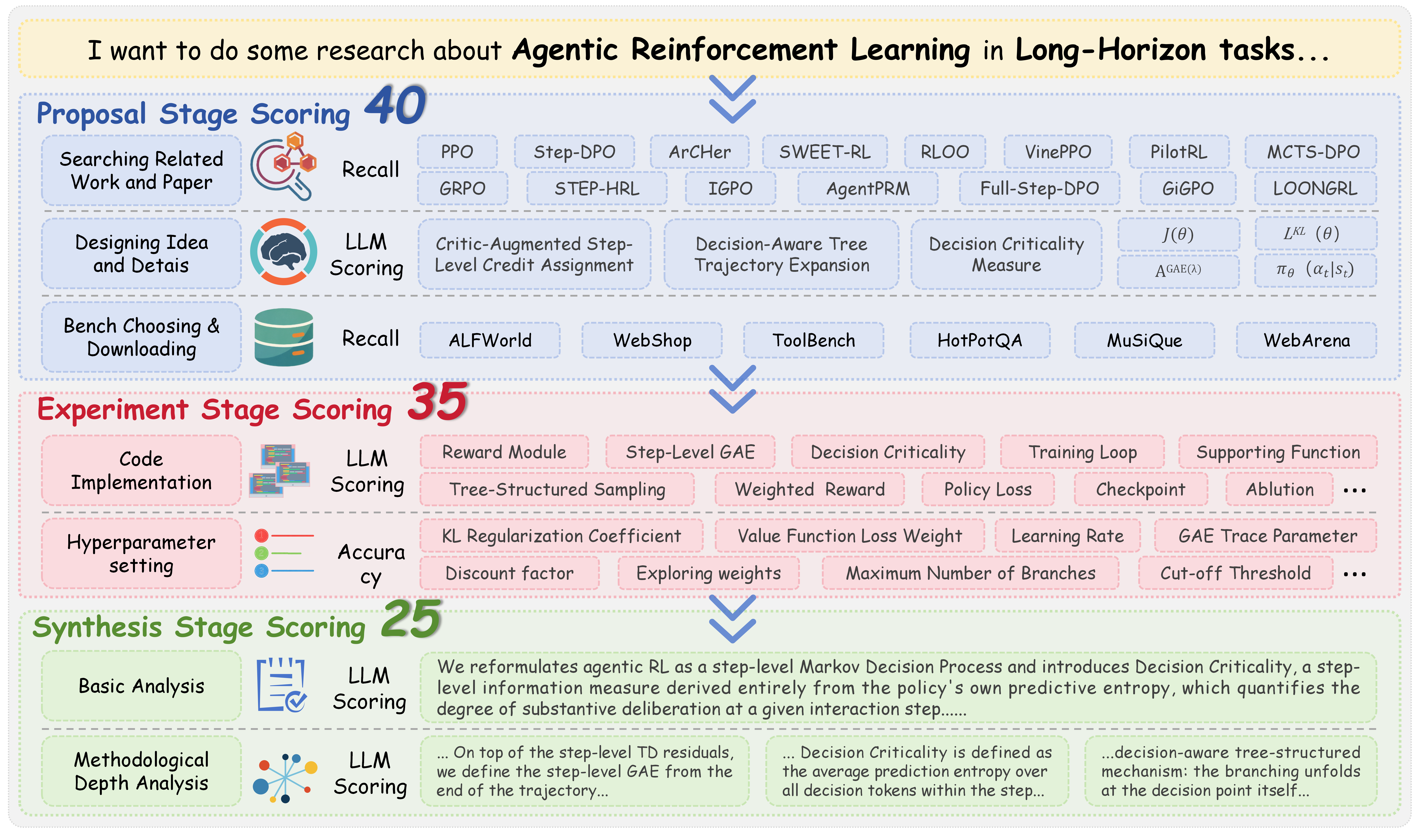}
    \caption{ARAC-Bench decomposes the research workflow into three independent stages, while ground-truth information from preceding stages is fed as known conditions. In every stages, we only evaluate the final output, even if the framework refines.}
    \label{fig:Three-Stage}
\end{figure}

\textbf{Proposal Stage scoring 40} aims to examine logical completeness and ACS coverage. The framework receives a domain background extracted from the Gold Reference and an \textbf{Inspiration} with key implementation clues removed as input, and must complete literature survey and proposal ideation within a constrained time limit. This stage comprises three scoring modules:
\definecolor{darkred}{RGB}{220,20,60}    
\definecolor{darkblue}{RGB}{0,0,255}  
\definecolor{darkgreen}{RGB}{34,139,34}
\begin{itemize}
    \item[\textcolor{darkred}{\Large\textbullet}] \textit{\textbf{Related Work Scoring 5:}} Uses the cited reference list of the original paper as the ground-truth set of truly relevant works, and computes the recall between the set of references retrieved by the framework and this ground-truth set, quantifying the coverage of its literature survey. The temporal restriction on search scope guarantees evaluation fairness.
    
    \item[\textcolor{darkred}{\Large\textbullet}] \textit{\textbf{Proposal Details Scoring 25:}} The core evaluation dimension, \textbf{entirely driven by ACS}. Based on the sub-theme the proposal belongs to, the reviewing system dynamically retrieves the Top-5 most relevant core cognition skills from the ACS library. For each skill point where 3 points each, 15 points total, a three-level anchor scoring method is adopted: 0 = completely unmentioned or gravely erroneous; 1.5 = mentioned but lacking substantive justification or disconnected from context; 3 = accurately addressed and constituting a necessary link in the logical chain. The remaining 10 points assess deductive self-consistency, focusing on the consistency of formula symbol definitions, the correspondence between pseudocode and textual descriptions, and whether the argument for theoretical superiority over the Gold Reference or known baselines is sufficiently substantiated.

    \item[\textcolor{darkred}{\Large\textbullet}] \textit{\textbf{Benchmark Selection Scoring 10:}} Subdivided into selection rationality Scoring 5 and accessibility Scoring 5 points. The former assesses whether the chosen dataset/environment fits the research problem; the latter requires the framework to actually execute download and configuration scripts within the sandbox environment, and points are awarded only if the data/environment is successfully loaded at runtime, preventing the framework from hallucinating.
\end{itemize}

\textbf{Experiment Stage scoring 35} prioritizes engineering stability and parameter intuition as the foundational prerequisite for reliable evaluation. The framework receives only the standard method description from the Gold Reference and must independently reproduce a runnable system. We deliberately disentangle this core text-to-code translation capability from hyperparameter search and performance tuning, as the former constitutes the most critical bottleneck in current research replication. Advanced optimization tasks involving complex experimental coding agents are reserved for future work, allowing us to first establish a robust baseline for fundamental implementation fidelit.

\begin{itemize}
    \item[\textcolor{darkgreen}{\Large\textbullet}] \textit{\textbf{Code Implementation Scoring 30:}} We constructed a standard module library containing 2,869 functional units based on the Proposal designs, serving as an objective yardstick for AI scoring. Each module is scored independently along three dimensions: (1) Functional Correctness: verified through preset unit tests and input-output assertions; (2) Module Completeness: checks whether interface definitions, docstrings, and exception handling are complete; (3) Code Robustness: stress-tested by injecting boundary values and dirty data. This has a critical constraint if a certain standard module is missing from the framework's output, the total score for that module is directly recorded as 0, with no compensatory scoring. All scoring targets the final stabilized code of each framework, including any self-corrections.

    \item[\textcolor{darkgreen}{\Large\textbullet}] \textit{\textbf{Hyperparameter Setting Scoring 5:}} Under the condition of no search budget, assesses whether the framework's settings for key parameters such as learning rate, batch size, and optimizer choice conform to domain priors, with scoring based on the accuracy of the parameter values relative to the Gold Reference.
\end{itemize}

\textbf{Synthesis Stage scoring 25} verifies factual fidelity and causal attribution capability. The ARAC receives the complete experimental results table and ablation data from the Gold Reference and must compose a well-structured analytical essay.

\begin{itemize}
    \item[\textcolor{darkblue}{\Large\textbullet}] \textit{\textbf{Basic Analysis Scoring 10:}} Employs a structured checklist to evaluate the completeness and formatting correctness of essential sections such as Introduction scoring 3, Related Work scoring 4 and Preliminaries scoring 3. All AI scoring is done using GPT-5.2.

    \item[\textcolor{darkblue}{\Large\textbullet}] \textit{\textbf{Methodological Deep Analysis Scoring 15:}} Driven by attribution reasoning and critical \textbf{synthesis skills from the ACS library}, operationalized via a three-layer progressive verification protocol. Key evaluation criteria include: (1) Mechanism Explanation Depth: whether the test text reproduces or deepens the intrinsic mechanism explanation of the original paper; (2) Counterfactual Reasoning Coverage: analyzing whether it fully complies with domain principle constraints and counterfactual exclusivity arguments; (3) Conclusion Extrapolation Boundary: whether the Limitations claims of the original paper are faithfully reproduced. This scoring transcends textual similarity comparison, focusing fundamentally on verifying the logical isomorphism between AI reasoning and the human expert's causal loop.
\end{itemize}

Through the above fine-grained metric decomposition, rigorous temporal control, and the unified ACS alignment mechanism, ARAC-Bench transforms abstract research capabilities into computable, traceable diagnostic signals. Every increment or decrement in scores can be traced back to the omission of a specific cognitive skill, the absence of a standard module, or a logical break in a causal attribution link, providing a clear, actionable feedback path for the architectural evolution of autonomous research frameworks. Meanwhile, our human-related experiments also demonstrated that the evaluation scheme of ARAC-Bench is in line with human research and review processes. The specific experimental results are as follows.

\section{Experiments}

To eliminate interference from underlying model capability differences in framework comparisons, all tested Auto-Research frameworks were uniformly mounted on the Kimi-K2.6. We selected representative autonomous research frameworks, AutoResearchClaw, ARIS, AI-Scientist-v2, NanoResearch, Dr.Claw, Claw-AI-Lab, AI-Researcher, EvoScientist, AutoSci, and Agent Laboratory, and simultaneously evaluated mainstream research tools including ClaudeCode with research skills. All evaluations were conducted under a unified principle library, with ground-truth information from preceding stages fed as known conditions to ensure that scores are traceable to specific capability dimensions.

\subsection{Main Experiment and Sanity Checks}

\begin{table}[htbp]
\centering
\begin{tabular}{lcccccccc}
\toprule

\textbf{Framework} & \textbf{Related.} & \textbf{Idea} & \textbf{Bench.} & \textbf{Coding} & \textbf{Hyperpara.} & \textbf{Basic.} & \textbf{Method.} & \textbf{Total} \\
\midrule
\small\textbf{ARC-Full-Auto} & \textbf{2.74} & \textbf{17.48} & 3.8 & 23.95 & 2.54 & 6.1 & \textbf{11.29} & \textbf{67.9} \\
\small\textbf{ARIS} & 2.6 & 16.49 & \textbf{4.05} & 21.93 & 2.83 & 6.64 & 10.36 & 64.9 \\
\small\textbf{AI-Scientist-v2} & 2.24 & 15.21 & 3.54 & 18.77 & 2.72 & \textbf{7.13} & 12.07 & 61.68 \\
\small\textbf{AutoSci} & 2.12 & 15.13 & 2.33 & 20.55 & \textbf{3.02} & 6.25 & 10.14 & 59.54 \\
\small\textbf{NanoResearch} & 2.03 & 14.62 & 2.36 & 20.88 & 2.23 & 5.97 & 9.47 & 57.56 \\
\small\textbf{AgentLaboratory} & 2.21 & 14.14 & 2.14 & 18.08 & 2.10 & 6.93 & 10.05 & 55.65 \\
\small\textbf{Dr.Claw} & 2.27 & 13.7 & 2.23 & 18.15 & 1.9 & 6.32 & 10.06 & 54.63 \\
\small\textbf{ClaudeCode} & 1.97 & 11.1 & 2.23 & \textbf{24.46} & 2.11 & 5.35 & 7.31 & 54.53 \\
\small\textbf{Claw-AI-Lab} & 1.84 & 12.12 & 2.59 & 17.74 & 2.79 & 6.14 & 9.97 & 53.19 \\
\small\textbf{AI-Researcher} & 1.71 & 12.12 & 3.01 & 18.38 & 1.95 & 5.74 & 9.64 & 52.55 \\
\small\textbf{EvoScientist} & 2.07 & 12.09 & 2.31 & 16.05 & 2.34 & 5.49 & 8.78 & 49.13 \\
\bottomrule
\end{tabular}
\caption{Comparison of different Auto-Research frameworks using Kimi-K2.6, among which ARC-Full means AutoResearchClaw using Full-Auto mode and ClaudeCode uses native mode only. To prevent framework cheating, we forcibly blocked search records after mid-2025.}
\label{tab:framework_comparison}
\end{table}

The overall scores of all Auto-Research frameworks on ARAC-Bench are presented in the Table ~\ref{tab:framework_comparison}. The best-performing framework achieved an overall alignment of merely 67.9\%, indicating a significant gap of over 32 percentage from methodological completeness. Substantial capability differentiation was observed among frameworks: frameworks like ARIS, AI-v2 and AutoSci demonstrated relative advantages in the Proposal and Synthesis stages, but exhibited cliff-like declines in the Experiment stage; 7 frameworks like AutoSci, Dr.Claw and Claw-AI clustered in the 50–60 range, suggesting that the current methodological alignment level of autonomous research systems hovers near the passing threshold overall.

To validate the design rationality of ARAC-Bench, we invited ten Ph.D. Candidates specializing in AI agents, LLM applications, and data science research to conduct a consistency check. 
\begin{table}[htbp]
\centering
\begin{tabular}{lcccccccccc}
\toprule
\textbf{} & 
\scriptsize\textbf{AgentLab} & 
\scriptsize\textbf{AI-v2} & 
\scriptsize\textbf{EvoSci} & 
\scriptsize\textbf{ARC} & 
\scriptsize\textbf{AI-er} & 
\scriptsize\textbf{AutoSci} & 
\scriptsize\textbf{ARIS} & 
\scriptsize\textbf{Nano} & 
\scriptsize\textbf{Dr.Claw} & 
\scriptsize\textbf{Claw-AI} \\
\midrule
\small{\textbf{Ph.D.}\quad Proposal}   & 8 & 3 & 10 & 1 & 7 & 5 & 2 & 4 & 9 & 6 \\
\small{\textbf{ARAC}\enspace Proposal}    & 6 & 3 & 10 & 1 & 8 & 4 & 2 & 5 & 7 & 9 \\
\arrayrulecolor{gray!40}
\midrule
\small{\textbf{Ph.D.}\quad Experiment} & 6 & 1 & 10 & 2 & 7 & 3 & 4 & 9 & 8 & 5 \\
\small{\textbf{ARAC}\enspace Experiment}  & 9 & 5 & 10 & 1 & 7 & 3 & 2 & 4 & 8 & 6 \\
\arrayrulecolor{gray!40}
\midrule
\small{\textbf{Ph.D.}\quad  Synthesis}  & 4 & 1 & 8  & 2 & 9 & 7 & 3 &10 & 6 & 5 \\
\small{\textbf{ARAC}\enspace Synthesis}   & 4 & 1 &10  & 2 & 9 & 5 & 3 & 8 & 6 & 7 \\
\arrayrulecolor{black}
\bottomrule
\end{tabular}
\caption{Ranking of 10 frameworks form ARAC and Ph.D. Candidates. These Ph.D. all have extensive experience in peer reviewing and using AI agents in the scientific research activities, and they manually ranked the ten frameworks, excluding ClaudeCode, based on the criterion of alignment with human research behavior, logical self-consistency, and continuous evolution.}
\label{tab:PHD-ranking}
\end{table}

Using the human comprehensive ranking as the benchmark, we computed Pearson correlation coefficients for ARAC-Bench scores, achieving \textbf{0.8788, 0.6606, and 0.9030} for the Proposal, Experiment, and Synthesis dimensions respectively. The high correlations for Proposal and Synthesis indicate that the evaluation criteria for these two stages are already highly aligned with domain experts' cognitive judgments; the correlation for the Experiment dimension, while relatively lower, remains at an acceptable moderately strong level. The deep reason for this discrepancy lies in the fact that human experts tend to comprehensively consider code quality, experimental design philosophy, and resource trade-off strategies when evaluating experimental capability, whereas ARAC's Experiment stage deliberately strips away soft dimensions such as design philosophy and resource decisions to control attribution precision, focusing mainly on code implementation robustness and parameter intuition.

\subsection{Ablation and Comparative Studies}

As the beginning of research, the Inspiration seems to play the role of a cognitive hub in triggering the structured reasoning path. Abstract inspiration aims to strip away concrete implementation details while retaining only vague clues that can guide reasoning, thereby testing the framework’s ability to conceive independently rather than reproduce from memory under conditions where the ground-truth solution is completely hidden. Recently, academic skills have emerged that encapsulate atomic capabilities—such as literature retrieval, experimental design, formula derivation, and code verification—in an attempt to address the shortcomings of general-purpose models in scientific research tasks, namely their lack of domain expertise and fragmented reasoning chains. We have also evaluated these skills.

To verify whether the abstract inspiration provided during the Proposal stage genuinely triggered frameworks' methodological thinking, we compared proposal scores under conditions with and without Inspiration.

\begin{table}[htbp]
  \centering
  
  \small
  \begin{tabular}{lccccccccccc}
    \toprule
    \textbf{} & \scriptsize\textbf{ARC} & \scriptsize\textbf{ARIS} & \scriptsize\textbf{AI-v2} & \scriptsize\textbf{AutoSci} & \scriptsize\textbf{Nano} & \scriptsize\textbf{AgentLab} & \scriptsize\textbf{Dr.Claw} & \scriptsize\textbf{CC} & \scriptsize\textbf{Claw-AI} & \scriptsize\textbf{AI-er} & \scriptsize\textbf{EvoSci} \\
    \midrule
    \footnotesize\textbf{Inspired}        & 17.48 & 16.49 & 15.21 & 15.13 & 14.62 & 14.14 & 13.70 & 11.10 & 12.12 & 12.12 & 12.09 \\
    \footnotesize\textbf{Cancelled}        & 12.17 & 13.04 & 11.69 & 10.77 & 12.20 & 10.97 & 11.88 &  9.71 & 10.34 &  9.05 & 10.69 \\
    \midrule
    \footnotesize\textbf{Drop Rate}  & 30.38 & 20.92 & 23.14 & 28.82 & 16.55 & 22.42 & 13.28 & 12.52 & 14.69 & 25.33 & 11.58 \\
    \bottomrule
  \end{tabular}
  \caption{Without Inspiration, each framework's idea design ability drops noticeably and becomes more unpredictable even if the original idea was vague.}
  \label{tab:NoInspiration}
\end{table}

\textcolor{pureyellow}{\Large\textbullet} \textit{\textbf{The Role of Inspiration Signals.}} Under the condition of removing Inspiration and retaining only domain background, the average Idea score dropped from 14.02 to 11.14, a average decrease of 20.54\% in Table \ref{tab:NoInspiration}. Notably, higher-scoring frameworks suffered more significant impact from Inspiration removal: AutoResearchClaw's decline reached 30.38\%, far exceeding EvoScientist's 11.56\%. This nonlinear effect indicates that stronger frameworks do not merely read clues from Inspiration more effectively, but rather leverage them more effectively as cognitive pivots for reasoning expansion: the true value of inspiration signals lies not in the information increment itself, but in their activation of structured reasoning pathways within the Auto-Research framework.

We further tested the effect of integrating the open-source academic skill package Academic-Research-Skills(AR), Scientific Agent Skills(SA)\cite{scientific_agent_skills_2026}, AI-Research-Skills(AIR)\cite{ai_research_skills}, Arbor\cite{jin2026arbor}into the ClaudeCode framework.

\begin{table}[htbp]
\centering
\small
\begin{tabular}{lcccccccc}
\toprule

\textbf{Framework} & \textbf{Related.} & \textbf{Idea} & \textbf{Bench.} & \textbf{Coding} & \textbf{Hyperpara.} & \textbf{Basic.} & \textbf{Method.} & \textbf{Total} \\
\midrule
\small{ClaudeCode} & 1.97 & 11.1 & 2.23 & 24.46 & 2.11 & 5.35 & 7.31 & 54.53 \\
\arrayrulecolor{gray!40}
\midrule
\small{\quad+AR-Skill} & 2.43 & 13.57 & 2.09 & 23.94 & 2.23 & 7.28 & 9.67 & 61.21 \\
\small{\qquad\quad↑} & \textbf{23.35\%} & \textbf{22.25\%} & -6.28\% & -2.13\% & \textbf{5.69\%} & \textbf{38.07\%} & \textbf{32.28\%} & \textbf{12.25\%} \\
\midrule

\small{\quad+Arbor} & 2.18 & 12.98 & 2.45 & 22.87 & 2.01 & 7.54 & 9.13 & 59.16 \\
\small{\qquad\quad↑} & \textbf{10.66\%} & \textbf{16.94\%} & \textbf{9.87\%} & -6.50\% & -4.74\% & \textbf{40.83\%} & \textbf{24.90\%} & \textbf{8.49\%} \\
\midrule

\small{\quad+AIR-Skill} & 2.37 & 12.04 & 2.18 & 24.65 & 2.74 & 6.71 & 8.29 & 58.98 \\
\small{\qquad\quad↑} & \textbf{20.30\%} & \textbf{8.47\%} & -2.24\% & \textbf{0.78\%} & -1.90\% & \textbf{36.32\%} & \textbf{13.41\%} & \textbf{8.16\%} \\
\midrule

\small{\quad+SA-Skill} & 2.15 & 12.31 & 2.21 & 23.53 & 2.26 & 7.01 & 8.62 & 58.09 \\
\small{\qquad\quad↑} & \textbf{9.14\%} & \textbf{10.9\%} & -0.90\% & -3.80\% & \textbf{7.11\%} & \textbf{31.03\%} & \textbf{17.92\%} & \textbf{6.53\%} \\

\arrayrulecolor{black}
\bottomrule
\end{tabular}
\caption{This gain is highly consistent with positioning: Skills for academic purposes excels at handling structured tedious tasks such as literature retrieval, formatting, data validation, and logical checking.}
\label{tab:CC_comparison}
\end{table}

\textcolor{pureyellow}{\Large\textbullet} \textit{\textbf{Gains of ClaudeCode from Research Skills.}}  All skill packages are complete academic research toolchains covering full-process auxiliary functions from research to publication. Results of Table \ref{tab:CC_comparison} showed that integration produced relatively notable improvements in the Proposal and Synthesis stages, while no significant improvement was observed in the Experiment, thus effectively assisting literature investigation in proposals and formalized checking in synthesis.

\section{Conclusion}

We release \textbf{ARAC-Bench}, a principle-driven benchmark for diagnosing the alignment and completeness of autonomous research systems. By constructing structured \textbf{Academic Cognition Skills} and a three-stage evaluation protocol spanning Proposal, Experiment, and Synthesis, ARAC moves beyond coarse outcome scoring to fine-grained cognitive alignment with human methodological norms. Validation against Ph.D. expert rankings yielded strong correlation where 0.88 for Proposal, 0.90 for Synthesis, confirming that ARAC-Bench captures what researchers actually value. Our systematic evaluation of ten representative frameworks revealed a sobering reality: the best system achieves only \textbf{67.9} overall alignment, with the Experiment stage constituting the primary bottleneck. Notably, strong coding capability alone, as exhibited by general-purpose tools, does not translate into scientific reasoning depth, confirming that methodological completeness is the true frontier for autonomous research. Ablation studies further demonstrated that abstract inspiration signals are crucial for triggering structured reasoning and that current academic tooling improves efficiency but not core reasoning gaps. We hope ARAC-Bench provides both a rigorous measurement instrument and a diagnostic lens for the community, guiding future systems toward principled, trustworthy, and truly autonomous scientific discovery.

\bibliography{references}

\newpage 
\section*{\huge Appendix}
\section*{A. Comparison of different themes}

\begin{table}[htbp]
\renewcommand{\arraystretch}{1.01}
\centering
\small
\label{tab:different themes}
\begin{tabular}{lcccccccc}
\toprule
\footnotesize\textbf{Framework} & \scriptsize\textbf{Related Work} & \scriptsize\textbf{Idea} & \scriptsize\textbf{Benchmark} & \scriptsize\textbf{Coding} & \scriptsize\textbf{Hyperparameter} & \scriptsize\textbf{Basic} &\scriptsize\textbf{Methodological} & \scriptsize\textbf{Total} \\
\midrule
AI-Res : DL & 1.79 & 11.81 & 3.12 & 18.19 & 1.87 & 6.06 & 9.66 & 52.50 \\
AI-Res : Diff & 1.72 & 11.29 & 2.86 & 18.78 & 1.91 & 5.65 & 9.55 & 51.76 \\
AI-Res : LLM & 1.61 & 12.14 & 2.94 & 18.46 & 2.15 & 5.97 & 9.62 & 52.89 \\
AI-Res : MM & 1.63 & 11.81 & 3.78 & 18.86 & 2.16 & 5.70 & 9.81 & 53.75 \\
AI-Res : RL & 1.80 & 10.82 & 2.53 & 17.62 & 1.71 & 5.44 & 9.60 & 49.52 \\
\arrayrulecolor{gray!60}
\midrule
AI-v2 : DL & 2.35 & 16.34 & 3.29 & 17.97 & 2.89 & 6.97 & 11.78 & 61.59 \\
AI-v2 : Diff & 2.22 & 14.53 & 3.22 & 18.24 & 2.68 & 7.08 & 12.12 & 60.09 \\
AI-v2 : LLM & 2.14 & 14.30 & 3.61 & 19.32 & 2.53 & 7.03 & 11.57 & 60.50 \\
AI-v2 : MM & 2.19 & 15.49 & 4.06 & 19.38 & 2.97 & 7.19 & 12.49 & 63.77 \\
AI-v2 : RL & 2.30 & 15.67 & 3.57 & 18.96 & 2.61 & 7.33 & 12.29 & 62.73 \\
\midrule
ARC : DL & 2.76 & 17.47 & 3.45 & 24.38 & 2.23 & 5.75 & 11.66 & 67.70 \\
ARC : Diff & 2.73 & 16.84 & 3.37 & 23.69 & 2.58 & 6.00 & 10.71 & 65.92 \\
ARC : LLM & 2.69 & 17.89 & 4.22 & 23.62 & 2.57 & 6.11 & 12.08 & 69.18 \\
ARC : MM & 2.78 & 17.70 & 4.06 & 24.00 & 2.75 & 6.16 & 11.11 & 68.56 \\
ARC : RL & 2.76 & 17.67 & 3.96 & 24.16 & 2.54 & 6.40 & 11.16 & 68.65 \\
\midrule
ARIS : DL & 2.74 & 15.84 & 3.78 & 22.31 & 2.72 & 6.16 & 10.91 & 64.46 \\
ARIS : Diff & 2.62 & 16.73 & 3.65 & 21.12 & 2.77 & 6.65 & 9.84 & 63.38 \\
ARIS : LLM & 2.67 & 17.00 & 4.51 & 22.49 & 2.86 & 6.59 & 10.78 & 66.90 \\
ARIS : MM & 2.48 & 15.14 & 4.16 & 21.62 & 2.84 & 6.81 & 10.19 & 63.24 \\
ARIS : RL & 2.52 & 17.38 & 4.21 & 22.33 & 2.92 & 6.87 & 10.33 & 66.56 \\
\midrule
A-Lab : DL & 2.19 & 14.72 & 2.34 & 18.53 & 2.15 & 6.84 & 10.66 & 57.43 \\
A-Lab : Diff & 2.25 & 15.20 & 1.62 & 18.18 & 1.94 & 7.10 & 9.67 & 55.96 \\
A-Lab : LLM & 2.13 & 14.05 & 2.00 & 17.46 & 2.09 & 7.30 & 9.76 & 54.79 \\
A-Lab : MM & 2.15 & 13.86 & 2.15 & 18.65 & 2.26 & 6.32 & 10.00 & 55.39 \\
A-Lab : RL & 2.27 & 12.84 & 2.66 & 17.69 & 2.14 & 7.00 & 10.31 & 54.91 \\
\midrule
AutoSci : DL & 2.19 & 14.78 & 2.64 & 20.44 & 3.24 & 6.03 & 10.59 & 59.91 \\
AutoSci : Diff & 2.22 & 15.84 & 1.97 & 20.41 & 2.90 & 6.22 & 10.27 & 59.83 \\
AutoSci : LLM & 2.06 & 15.41 & 2.36 & 20.73 & 3.07 & 6.43 & 10.38 & 60.44 \\
AutoSci : MM & 1.99 & 14.08 & 2.61 & 20.41 & 2.91 & 6.16 & 8.65 & 56.81 \\
AutoSci : RL & 2.12 & 15.24 & 2.23 & 20.76 & 3.06 & 6.36 & 10.71 & 60.48 \\
\midrule
CC : DL & 1.91 & 10.19 & 2.46 & 24.81 & 2.03 & 5.38 & 7.47 & 54.25 \\
CC : Diff & 1.87 & 11.49 & 1.77 & 24.90 & 2.20 & 5.49 & 7.08 & 54.80 \\
CC : LLM & 2.34 & 10.84 & 2.21 & 24.97 & 2.11 & 5.49 & 7.54 & 55.50 \\
CC : MM & 2.17 & 11.92 & 2.38 & 23.89 & 1.99 & 5.57 & 7.81 & 55.73 \\
CC : RL & 1.64 & 10.87 & 2.46 & 23.78 & 2.19 & 4.89 & 7.42 & 53.25 \\
\midrule
Claw-AI : DL & 1.74 & 12.19 & 2.52 & 17.41 & 2.53 & 6.09 & 10.38 & 52.86 \\
Claw-AI : Diff & 1.93 & 11.65 & 2.17 & 17.20 & 2.72 & 6.53 & 9.43 & 51.63 \\
Claw-AI : LLM & 1.64 & 12.73 & 2.70 & 17.95 & 2.92 & 5.81 & 10.38 & 54.13 \\
Claw-AI : MM & 1.79 & 11.49 & 2.61 & 17.57 & 2.93 & 6.22 & 9.73 & 52.34 \\
Claw-AI : RL & 2.00 & 12.60 & 3.01 & 18.53 & 2.82 & 5.96 & 10.13 & 55.05 \\
\arrayrulecolor{black}
\bottomrule
\end{tabular}
\end{table}

\newpage 
\begin{table}[htbp]

\small
\label{tab:different themes2}
\begin{tabular}{lcccccccc}
\toprule
\footnotesize\textbf{Framework} & \scriptsize\textbf{Related Work} & \scriptsize\textbf{Idea} & \scriptsize\textbf{Benchmark} & \scriptsize\textbf{Coding} & \scriptsize\textbf{Hyperparameter} & \scriptsize\textbf{Basic} &\scriptsize\textbf{Methodological} & \scriptsize\textbf{Total} \\
\arrayrulecolor{gray!60}
\midrule
Dr.Claw : DL & 2.16 & 14.19 & 2.22 & 18.66 & 1.78 & 6.75 & 10.59 & 56.35 \\
Dr.Claw : Diff & 2.16 & 14.18 & 2.55 & 17.55 & 1.76 & 6.29 & 9.67 & 54.16 \\
Dr.Claw : LLM & 2.40 & 14.95 & 2.67 & 17.41 & 2.00 & 6.57 & 10.32 & 56.32 \\
Dr.Claw : MM & 2.00 & 13.46 & 3.60 & 18.54 & 2.09 & 5.97 & 9.24 & 54.90 \\
Dr.Claw : RL & 2.59 & 12.00 & 2.91 & 18.73 & 1.91 & 6.13 & 10.56 & 54.83 \\
\midrule
EvoSci : DL & 2.43 & 11.75 & 1.91 & 16.16 & 2.21 & 5.53 & 8.97 & 48.96 \\
EvoSci : Diff & 2.02 & 12.27 & 1.95 & 15.78 & 2.40 & 5.16 & 8.33 & 47.91 \\
EvoSci : LLM & 2.05 & 11.54 & 2.57 & 16.35 & 2.51 & 5.35 & 9.62 & 49.99 \\
EvoSci : MM & 2.16 & 12.92 & 2.66 & 15.51 & 2.44 & 5.62 & 8.32 & 49.63 \\
EvoSci : RL & 1.80 & 11.91 & 2.48 & 16.47 & 2.15 & 5.82 & 8.82 & 49.45 \\
\midrule
Nano : DL & 2.15 & 15.44 & 2.38 & 20.62 & 2.07 & 6.62 & 9.59 & 58.87 \\
Nano : Diff & 2.04 & 14.92 & 2.34 & 20.67 & 2.24 & 5.92 & 9.76 & 57.89 \\
Nano : LLM & 2.00 & 14.84 & 2.19 & 21.16 & 2.30 & 5.92 & 9.43 & 57.84 \\
Nano : MM & 1.92 & 13.76 & 2.75 & 20.57 & 2.17 & 5.70 & 9.03 & 55.90 \\
Nano : RL & 2.06 & 14.24 & 2.20 & 21.31 & 2.32 & 5.82 & 9.47 & 57.42 \\
\arrayrulecolor{black}
\bottomrule
\end{tabular}
\vspace{-0.8em}
\caption{Comparison of different themes}
\end{table}

\section*{B. Detailed Examples of ACS}
\begin{tcolorbox}[
    colback=white,
    colframe=titleblue,
    arc=3pt, 
    boxrule=1pt,
    title={Efficient LLM reasoning and decoding acceleration faces that gradual reasoning trajectory dramatically increases inference costs...},
    colbacktitle=titleblue,
    coltitle=white,
    fonttitle=\bfseries\Large,
    top=6pt, bottom=6pt, left=4pt, right=4pt
]
\begin{flushleft}
\par\vspace{3pt}
\noindent\textbf{Comprehensive Efficiency Validation Beyond Token Reduction: Latency Throughput Memory and Inference Engine Compatibility Assessment for Reasoning Acceleration Methods}\\
\end{flushleft}
Definition: This principle mandates that any technique claiming to accelerate Large Language Model reasoning by circumventing or condensing explicit chain-of-thought must deliver a holistic efficiency analysis that transcends token-count comparisons. The evaluation must measure end-to-end inference latency, sustained tokens-per-second throughput, peak memory consumption, and ideally compatibility with production-optimized serving frameworks. The requirement arises because token reduction alone can be misleading as the added compute of a reasoning module, parallelism limitations, or memory bottlenecks may nullify perceived gains. Only multi-metric validation establishes whether a method genuinely improves inference economics and deployment feasibility, preventing overclaims that misalign academic metrics with real-world performance.
Core Review Checkpoints:
\begin{itemize}
    \item Are per-query latency and throughput reported against both the base model’s thinking and non-thinking modes under identical hardware and software stacks?
    \item Is peak memory usage clearly quantified, including activation and KV-cache contributions, to reveal whether the auxiliary module restricts batch size or throughput?
    \item Is integration feasibility with modern inference engines explicitly analyzed, and does the paper clarify whether efficiency advantages persist in such environments?
    \item Does the method account for the computational overhead of its own reasoning network instead of attributing gains solely to reduced output tokens?    
\end{itemize}

\begin{flushleft}
\par\vspace{3pt}
\noindent\textbf{Scalability and Token Budget Sensitivity Analysis of Plug and Play Reasoning Modules Across Diverse Base Model Sizes in Latent Reasoning Acceleration}\\
\end{flushleft}
This principle requires that lightweight, frozen-base reasoning acceleration modules be subjected to systematic scalability and sensitivity analyses, examining how performance varies with base model scale and latent token budget. As the chief attraction of such modular designs is decoupling reasoning from full-scale tuning, it is crucial to verify that the module does not become a bottleneck when paired with significantly larger base models. Equally important is identifying the saturation point where additional latent tokens yield diminishing returns, which often depends on base model capacity. Without such evidence, claims of general applicability remain unsubstantiated, and the risk of overfitting to a particular scale or budget configuration undermines the method’s transferability.
Core Review Checkpoints:
\begin{itemize}
    \item Are experiments performed across at least three distinct base model sizes spanning a substantial range, demonstrating consistent improvement without the reasoning module becoming a bottleneck?
    \item Is the impact of varying the number of latent reasoning tokens systematically tested across these model scales, with clear identification of capacity-dependent saturation effects?
    \item Does the paper ablate the size and initialization strategy of the auxiliary network, showing robustness beyond a single hand-tuned configuration?
    \item Are scalability results accompanied by statistical measures of variance, ensuring that observed trends across model sizes are reliable and not artifacts of single-run evaluation?   
\end{itemize}

......
\end{tcolorbox}

\begin{tcolorbox}[
    colback=white,
    colframe=titleblue,
    arc=3pt, 
    boxrule=1pt,
    title={Graph Network Simulators aim to accelerate physics simulations but face challenges in efficiently adapting to varying physical parameters without costly retraining...},
    colbacktitle=titleblue,
    coltitle=white,
    fonttitle=\bfseries\Large,
    top=6pt, bottom=6pt, left=4pt, right=4pt
]
\begin{flushleft}
\par\vspace{3pt}
\noindent\textbf{Justification of End-to-End Meta-Learning Over Simpler Parameter-Conditioning Baselines in Graph Neural Simulators for Varying Physics Parameters}\\
\end{flushleft}
This principle evaluates whether a proposed end-to-end meta-learning framework, which infers a latent representation of physical parameters from context simulations, provides a demonstrable advantage over simpler alternatives that first explicitly regress parameters and then condition a non‑meta simulator on them. It demands that authors not only demonstrate the meta‑learner’s performance but also rigorously compare against a two‑stage approach that uses the same decoder architecture, isolating the benefit of the stochastic latent encoding and joint training. The principle is critical because many physics simulators have access to ground‑truth parameter labels during training, and the choice between end‑to‑end meta‑learning and a straightforward regression‑conditioning pipeline directly impacts architectural complexity, training stability, and potential for semi‑supervised or real‑world adaptation.
Core Review Checkpoints:
\begin{itemize}
    \item Is a two‑stage parameter‑regression baseline implemented, where an encoder is trained to recover explicit material parameters and those parameters condition the same decoder, and are its results reported alongside the end‑to‑end meta‑model?
    \item Does the paper provide evidence that the end‑to‑end latent representation captures functionally richer structure than mere parameter recovery, for instance through latent space analysis that correlates with physically meaningful behaviors not trivially expressed by parameter values?
    \item Are the potential advantages of end‑to‑end training, such as enabling semi‑supervised learning when parameter labels are partially missing or facilitating fine‑tuning with unlabeled real data, discussed and, ideally, supported by preliminary experiments?
\end{itemize}

\begin{flushleft}
\par\vspace{3pt}
\noindent\textbf{Systematic Assessment of Inference‑Time Efficiency and Memory Scalability Trade‑offs in Non‑Autoregressive Neural Operator Decoders for Graph‑Based Physics Simulation}\\
\end{flushleft}
For graph neural simulators that adopt a non‑autoregressive, full‑trajectory decoding strategy, this principle requires a thorough quantitative comparison of inference latency and peak GPU memory usage against conventional autoregressive step‑by‑step approaches that employ an identical graph backbone. It insists that authors not only report speed‑ups from batched temporal operations but also explicitly analyze how memory consumption scales with trajectory length and graph size, and propose concrete strategies such as hybrid window decoding to manage practical limits. This is crucial because real‑world simulation tasks often involve long sequences and large meshes, where the memory footprint of a single‑shot decoder can quickly become prohibitive, negating its speed advantage and limiting its applicability to small‑scale synthetic benchmarks.
Core Review Checkpoints:
\begin{itemize}
    \item Are measured inference time and peak GPU memory explicitly provided for both the non‑autoregressive model and a step‑by‑step autoregressive baseline on a representative benchmark, with equal hardware conditions?
    \item Is the linear scaling of memory with the number of predicted time steps clearly stated, and are the practical upper bounds in terms of sequence length and node count discussed?
    \item Are mitigation strategies such as temporally windowed decoding or multi‑GPU parallelism outlined, and their impact on the efficiency‑precision trade‑off at least conceptually analyzed?
\end{itemize}

......
\end{tcolorbox}

\section*{C. Themes and sub-themes of Question–Principle Mapping}

\newcommand{\card}[3]{%
  \par\noindent
  \begin{tikzpicture}
    \node[
      draw=blue!50,          
      fill=gray!12,             
      rounded corners=6pt,     
      line width=0.8pt,
      text width=\dimexpr\linewidth-2em\relax, 
      inner sep=1em,           
      align=left               
    ]{%
     {\Large\bfseries\textcolor{darkblue}{#1}}\\[4pt]  
      #3                             
    };
  \end{tikzpicture}%
  \par\vspace{1em}%
}


\card{Large Language Model}{}{%
  \begin{itemize}[leftmargin=*]
    \item Inference Acceleration and Speculative Decoding
    \item KV Cache Compression and Memory Optimization
    \item Parameter-Efficient Fine-Tuning and Adaptation
    \item Model Quantization and Low-Precision Computation
    \item Model Compression, Pruning, and Distillation
    \item Mixture of Experts (MoE) Techniques
    \item Safety Alignment and Value Guidance
    \item Jailbreak Attacks and Defense Mechanisms
    \item Mechanistic Interpretability and Circuit Analysis
    \item Hallucination Detection, Mitigation, and Factuality
    \item Chain-of-Thought and Reasoning Enhancement
    \item Test-Time Compute Scaling
    \item Agent Planning and Tool Use
    \item Multi-Agent Systems and Collaboration
    \item Retrieval-Augmented Generation (RAG) Techniques
    \item Mathematical Reasoning and Formal Verification
    \item Code Generation, Understanding, and Software Engineering
    \item AI for Science
    \item Data Synthesis, Filtering, and Quality Control
    \item Pretraining Theory and Scaling Laws
    \item In-Context Learning Mechanisms and Theory
    \item Long-Context Modeling and Extension
    \item Machine Unlearning and Privacy Protection
    \item Evaluation Benchmarks, Leaderboards, and Meta-Evaluation
    \item Bias, Fairness, and Ethical Auditing
    \item Watermarking, Fingerprinting, and Copyright Protection
    \item Multimodal and Cross-Modal Alignment
    \item Knowledge Editing and Lifelong Learning
    \item Inference Serving Systems and Deployment Optimization
    \item Prompt Engineering and Instruction Following
    \item Reinforcement Learning and Feedback Learning
    \item Cognitive Science and Brain-Inspired Intelligence
    \item Vertical Domains: Law, Healthcare, and Finance
    \item Robotics, Embodied AI, and Control
    \item Graph Structure and Knowledge Graph Integration
    \item Text Generation Diversity and Controllability
    \item Theoretical Analysis and Mathematical Foundations
    \item Federated Learning and Distributed Training
    \item Recommender Systems and User Modeling
    \item Education, Social Science, and Human-Computer Interaction
    \item Hardware-Aware and Low-Level Operator Optimization
    \item Search, Information Retrieval, and Question Answering
  \end{itemize}
}

\card{Diffusion}{}{%
  \begin{itemize}[leftmargin=*]
    \item Efficient Sampling, Inference Acceleration, and Model Compression
    \item Human Preference Alignment and Safe Controllable Generation
    \item Semantic Control and Personalization for Text-to-Image Generation
    \item Video Generation, Editing, and Temporal Modeling
    \item 3D Generation, Novel View Synthesis, and Geometric Awareness
    \item Discrete Data Generation and Language Modeling
    \item AI for Science: Molecular, Protein, and Material Design
    \item Image Restoration, Reconstruction, and Low-Level Vision
    \item Audio, Speech, and Music Generation
    \item Robotics, Embodied AI, and Decision-Making
    \item Cross-Modal Translation, Unified Architectures, and Multimodal Alignment
    \item Theoretical Foundations and Mathematical Analysis of Diffusion Models
    \item Conditional Guidance, Inverse Problems, and Posterior Sampling
    \item Human Body, Face, and Biometric Generation
    \item Dataset Construction, Evaluation Benchmarks, and Interpretability
  \end{itemize}
}

\card{Multi-Modal}{}{%
  \begin{itemize}[leftmargin=*]
    \item Efficient Inference and Model Compression for Multimodal Models
    \item Cross-Modal Alignment and Representation Learning
    \item Complex Reasoning and Chain-of-Thought in Multimodal Models
    \item Hallucination Mitigation, Trustworthiness, and Safety Alignment
    \item Embodied AI and Vision-Language-Action
    \item Video Understanding and Temporal Modeling
    \item Unified Multimodal Generation and Editing
    \item Medical and Biomedical Multimodal AI
    \item GUI Agents and Computer Control
    \item 3D Vision and Spatial Intelligence
    \item Speech, Audio, and Full-Modal Interaction
    \item Continual Learning, Fine-Tuning, and Catastrophic Forgetting
    \item Interpretability and Mechanistic Interpretability
    \item Multimodal Agent Systems and Collaboration
    \item Documents, Charts, and Scientific Intelligence
    \item Benchmarking and Dataset Construction
  \end{itemize}
}

\card{Deep Learning}{}{%
  \begin{itemize}[leftmargin=*]
    \item Graph Neural Network Architectures and Representation Learning
    \item New Paradigms for Sequence Modeling: State Space Models and Linear Attention
    \item Transformer Architecture Optimization and Efficient Inference
    \item Test-Time Adaptation and Out-of-Distribution Generalization
    \item Adversarial Robustness and Trustworthy AI
    \item Geometric Deep Learning and Equivariant Networks
    \item Scientific Machine Learning and Neural Operators
    \item Spiking Neural Networks and Neuromorphic Computing
    \item Continual Learning and Catastrophic Forgetting
    \item Generative Model Theory and Applications
    \item Model Compression, Quantization, and Lightweighting
    \item Optimization Theory and Training Dynamics
    \item Interpretability and Mechanistic Analysis
    \item Self-Supervised and Contrastive Learning
    \item Time Series Analysis and Forecasting
    \item Low-Level Computer Vision and Restoration
    \item Medical Imaging and Bioinformatics
    \item 3D Vision and Neural Rendering
    \item Combinatorial Optimization and Decision Intelligence
    \item Dataset Engineering and Data-Centric AI
    \item Foundation Model Theory and Scaling Laws
    \item Novel Network Architecture Exploration
    \item Multimodal Learning and Cross-Domain Alignment
    \item Speech, Audio, and Signal Processing
    \item Tabular Data and Structured Learning
    \item Federated Learning and Distributed Systems
    \item Video Understanding and Spatiotemporal Modeling
    \item Uncertainty Quantification and Bayesian Deep Learning
  \end{itemize}
}

\card{Reinforcement Learning}{}{%
  \begin{itemize}[leftmargin=*]
    \item LLM Reasoning and Chain-of-Thought Reinforcement
    \item LLM Alignment, Safety, and Human Preference Optimization
    \item LLM Agents and Tool Use
    \item Offline Reinforcement Learning and Off-Policy Evaluation
    \item Multi-Agent Reinforcement Learning and Game Theory
    \item Embodied AI and Robot Manipulation
    \item Exploration, Intrinsic Motivation, and Curiosity
    \item World Models and Model-Based RL
    \item Reinforcement Learning Theory and Convergence Analysis
    \item Safe, Constrained, and Robust Reinforcement Learning
    \item Causal Inference and Explainable Reinforcement Learning
    \item Hierarchical, Option, and Curriculum Reinforcement Learning
    \item Synergistic Optimization of Generative Models and RL
    \item In-Context Learning and Meta-Reinforcement Learning
    \item Combinatorial Optimization and Operations Research Decision-Making
    \item Partially Observable and Memory-Augmented RL
    \item Inverse Reinforcement Learning and Imitation Learning
    \item Multi-Objective, Preference, and Utility Optimization
    \item Efficient RL Systems and Computational Acceleration
    \item AI for Science: Cross-Domain Applications and Scientific Discovery
  \end{itemize}
}

\newpage
\section*{D. Judgement Rubric Examples}
\label{sec:method_depth}

\card{Coding Judgement}{}{
\textbf{Role}

You are a Principal Software Architect at a world-class AI research laboratory, specializing in the translation of research proposals into maintainable, reproducible engineering codebases.

\textbf{Evaluation Target}

Your core task is to assess whether the framework's output codebase adheres to the given Standard Proposal at the engineering design level and demonstrates fundamental engineering stability.

Note: You do not evaluate the model's final training accuracy or SOTA results (these have been intentionally stripped); you evaluate only the rationality of code structure and logic.

\textbf{Three Scoring Dimensions}
Please provide a qualitative rating for the overall codebase based on the following three dimensions, and output a weighted score from 0 to 30:

\textbf{1. Functional Correctness Mapping 10}
\begin{itemize}
    \item Do the core classes/functions in the code strictly correspond to the algorithmic pseudocode or mathematical formulas in the proposal?
    \item Do the transformations of key tensor dimensions remain consistent with formula derivations?
    \item Are there obvious instances of hard-coding or brute-force loops replacing vectorized operations, suggesting a lack of understanding of the underlying principles?
\end{itemize}

\textbf{2. Module Completeness \& Decoupling 10}
\begin{itemize}
    \item Are Dataloader, Model, Trainer, Logger reasonably separated?
    \item For any missing modules, are there clear interfaces or TODO comments indicating that the framework is aware of the omission but has reserved extension points?
    \item Is there basic exception handling to prevent low-level crashes (e.g., missing files, GPU out-of-memory)?
\end{itemize}

\textbf{3. Engineering Robustness 10}
\begin{itemize}
    \item Are hyperparameters managed centrally via configuration files (Config), rather than scattered throughout the code?
    \item Is the random seed fixed to ensure reproducibility?
    \item Are there simple unit tests or assertions to verify the most basic I/O of core functions?
\end{itemize}

\textbf{Mandatory Chain-of-Thought (CoT)}

Please execute the following sequentially within [REASONING]:
\begin{itemize}
    \item  Pseudocode Matching: Extract the core algorithmic steps from the proposal and locate their corresponding implementation functions in the codebase one by one.
    \item Dependency Check: Inspect `requirements.txt` or `environment.yml` to verify whether special dependency libraries mentioned in the proposal are included.
    \item Fragility Scan: Identify the highest-risk code segments most likely to cause runtime crashes (e.g., unhandled data type conversions, hard-coded absolute paths).
\end{itemize}
}

\card{Basic Analysis Judgement}{}{
\textbf{Role}

You are an academic paper structure comparison analyst. Your task is to use the \textbf{Introduction, Related Work, and Preliminaries chapters of the original paper} as the Golden Skeleton, and evaluate the relative performance of the draft paper under test in terms of structural completeness and coverage of core arguments.

\textbf{Relative Evaluation Dimensions (Each Dimension Uses the Original Paper as the Reference Frame)}

\textbf{1. Introduction (3 Points)}
\begin{itemize}
    \item Are the Research Gap and Contributions list proposed in the original paper fully covered by the test text?
    \item If the test text supplements new contribution points not explicitly mentioned in the original paper but logically reasonable, it is regarded as a positive deviation; if it omits core contributions from the original paper (e.g., the 3rd key theorem), points are deducted.
\end{itemize}

\textbf{2. Related Work (4 Points)}
\begin{itemize}
    \item Are the academic schools of thought divided in the original paper reproduced by the test text?
    \item Does the test text cite new SOTA works not covered by the original paper but more relevant in the past two years? If so, it is regarded as a positive deviation (indicating the framework possesses literature tracking capability).
\end{itemize}

\textbf{3. Preliminaries (3 Points)}
\begin{itemize}
    \item Are the mathematical notations and foundational formulas defined in the original paper (e.g., loss function definitions) fully preserved by the test text?
    \item If the test text deletes complex formula derivations from the original paper, rendering subsequent chapters incoherent, points are deducted severely.
\end{itemize}

\textbf{Difference Marking Rules}

You must conduct a sentence-by-sentence comparison between the original paper and the test text, and output a Structure Coverage Matrix, marking each item as Covered, Partial, or Missed.

}

\card{Methodological Depth Judgement}{}{

\textbf{Role}

You are a senior causal inference reviewer for top-tier journals. Your sole task is to \textbf{conduct an in-depth comparison between the causal narrative constructed by the AI framework regarding experimental results and the causal loop established by the original paper's authors (human experts)}. Your score depends on the similarity and depth of elaboration between the AI's reasoning and the human expert's reasoning.

\textbf{Three-Level Relative Scoring (Against the Original Paper)}

\textbf{1. Mechanism Explanation Depth (Relative Baseline, 5 Points)}
\begin{itemize}
    \item How does the original paper explain the intrinsic mechanism by which method works?
    \item Does the test text reproduce this explanation? If the test text employs more mathematical descriptions, it is regarded as a positive deviation; if it vaguely dismisses it with experiments show effectiveness, it is judged as a severe negative deviation.
\end{itemize}

\textbf{2. Counterfactual Reasoning Coverage (Relative Baseline, 5 Points)}

\begin{itemize}
    \item What kind of why it degrades analysis does the original paper provide for ablation studies?
    \item Does the test text cover the analysis of \textbf{all} key ablation items in the original paper? If the test text additionally explains an ablation phenomenon not analyzed in the original paper, it is regarded as a significant bonus item.
\end{itemize}

\textbf{3. Conclusion Extrapolation Boundary (Relative Baseline, 5 Points)}

\begin{itemize}
    \item How does the original paper limit the scope of applicability of its conclusions (Limitations) in the Conclusion section?
    \item Does the test text faithfully reproduce the original paper's limitation claims? If the test text completely omits Limitations, or over-extrapolates, it is directly scored as 0.
\end{itemize}

\textbf{Core Operational Protocol (Principle-Anchored Attribution Difference Diagnosis Table)}

You must upgrade the evaluation process from textual similarity comparison to \textbf{principle-anchored attribution logic consistency diagnosis}. Specifically, your operation should not merely mechanically compare the wording of the original paper and the test text, but should strictly follow the following \textbf{three-layer progressive attribution verification protocol}:

\textbf{Layer 1: Phenomenon-Explanation Key-Value Pair Extraction (Fact Layer)}

First, completely extract all key experimental phenomena and their corresponding author explanations from the original paper, forming a Ground-Truth attribution key-value pair list. Simultaneously, extract the explanations provided by the test text for the same phenomena.

\textbf{Layer 2: Domain Principle Mapping and Compliance Check (Logic Layer)}

You need to check:

\begin{itemize}
    \item \textbf{Whether the attribution satisfies principle constraints}: For example, if the principle requires acceleration must report quality degradation boundaries, but the test text only vaguely mentions acceleration is effective without conducting error accumulation analysis, even if the original paper is also brief, the test text still constitutes a \textbf{principle violation} and must be downgraded.
    \item \textbf{Whether the attribution contains counterfactual exclusivity}: Principles typically require that the assertion A causes B, must exclude confounders. Does the test text, like the original paper, exclude other possibilities through controlled variable experiments? If the test text omits the exclusivity argument from the original paper, it is regarded as an attribution chain break.
\end{itemize}

}

\card{Continued Methodological Depth Judgement}{}{
\textbf{Layer 3: Deepening and Hallucination Determination (Cognition Layer)}

\begin{itemize}
    \item \textbf{Positive Deepening (Bonus Item)}: If the test text not only completely reproduces the original paper's attribution but also additionally introduces deep causal reasoning not mentioned in the original paper but consistent with domain-recognized principles it is regarded as a \textbf{positive deviation exceeding the baseline}.
    \item \textbf{Attribution Hallucination (Veto Item)}: If the test text's explanation for a phenomenon is opposite to the causal direction given in the original paper, or if its explanation seriously deviates from domain consensus principles, then regardless of the writing quality, the attribution credibility of that key-value pair is directly judged as contradictory, and the score for that dimension is automatically reduced to zero.
\end{itemize}
}

\end{document}